\documentclass{article}
\usepackage{amssymb}
\usepackage{graphicx}
\usepackage{xspace}
\usepackage{subcaption} % For subfigures
\usepackage{booktabs}

\usepackage[preprint]{corl_2025} % Uncomment for pre-prints (e.g., arxiv); This is like ``final'', but will remove the CORL footnote.

\newcommand{\ourvla}{Oat-VLA\xspace}
\newcommand{\libero}{LIBERO\xspace}
\newcommand{\dinosiglip}{DinoV2+SigLIP\xspace}

\newcommand{\oxe}{Open X-Embodiment\xspace}
\newcommand{\xarm}{UFACTORY xArm 6\xspace}

\title{Focusing on What Matters: \\ Object-Agent-centric Tokenization\\ for Vision Language Action Models}

\usepackage{amsmath}
\usepackage{subcaption}
\usepackage{enumitem}
\usepackage{placeins}

\usepackage[dvipsnames]{xcolor}
\usepackage{setspace}

\author{
  Rokas Bendikas\thanks{equal contribution}~~\thanks{work performed during internship at Qualcomm AI Research}\\
  Centre for Artificial Intelligence, UCL \\
  \texttt{r.bendikas@ucl.ac.uk} \\
  \And
  Daniel Dijkman\textsuperscript{\textasteriskcentered}\\
  Qualcomm AI Research\thanks{Qualcomm AI Research is an initiative of Qualcomm Technologies, Inc.}\\
  \texttt{ddijkman@qti.qualcomm.com} \\
  \And
  Markus Peschl \\
  Qualcomm AI Research \\
  \And
  Sanjay Haresh \\
  Qualcomm AI Research \\
  \And
  Pietro Mazzaglia \\
  Qualcomm AI Research \\
}

\begin{document}
\maketitle

\vspace{-2.5em}

\begin{abstract}

Vision-Language-Action (VLA) models offer a pivotal approach to learning robotic manipulation at scale by repurposing large pre-trained Vision-Language-Models (VLM) to output robotic actions. However, adapting VLMs for robotic domains comes with a high computational cost, which we attribute to the tokenization scheme of visual inputs.
In this work, we aim to enable more efficient VLA training by proposing \ourvla, an Object-Agent-centric Tokenization for VLAs. 
Building on the insights of object-centric representation learning, our method introduces an inductive bias towards scene objects
and the agent's own visual information.
As a result, we find that \ourvla can drastically reduce the number of visual tokens % by the agent
to just a few tokens without sacrificing performance. 
We reveal that \ourvla converges at least twice as fast as OpenVLA on the LIBERO suite, as well as outperforms OpenVLA in diverse real-world pick and place tasks.
\end{abstract}


\vspace{-1em}

\begin{figure}[h]
    \centering
    \begin{minipage}[t]{0.4\textwidth}
        \centering
        \includegraphics[width=\textwidth]{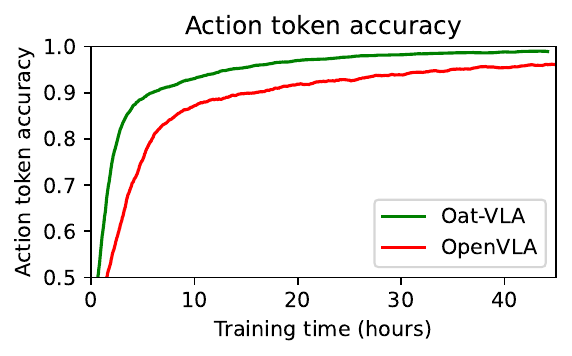}
    \end{minipage}
    \quad
    \begin{minipage}[t]{0.415\textwidth}
        \centering
        \includegraphics[width=\textwidth]{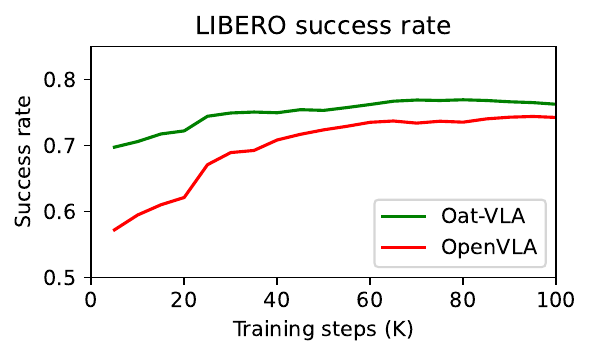}
    \end{minipage}
\caption{Comparison of our method \ourvla to OpenVLA on the \libero dataset, full fine-tuning.  \textbf{Left}: Action token accuracy relative to training time. \ourvla converges more than {\textcolor{black}{$2\times$ faster}}.
\textbf{Right}: Average success rate across the four \libero tasks suites relative to training steps. 
}
\label{fig:training_curves}
\end{figure}

\vspace{-1em}

\section{Introduction}
\label{sec:introduction}

Large language models (LLMs) \cite{vaswani2023attention, brown2020gpt2} have made it possible to tackle increasingly difficult tasks by leveraging self-supervised learning on large-scale datasets. Furthermore, the combination of the vast amount of knowledge present in modern pre-trained LLMs and the large amount of parameters have opened up many possibilities for adapting LLMs to a variety of applications, such as coding \cite{deepseekai2025r1}, reinforcement learning \cite{klissarov2023motif} and imitation learning \citep{brohan2023rt2}. A key ingredient thereof lies in the appropriate tokenization of the input data, which can be straightforward for natural language~\citep{kudo2018sentencepiece}, but is generally significantly more powerful when leveraging the structure of the input~\citep{sennrich2015neural}. While the latter has become a common strategy for language-valued inputs, the former choice of a simple tokenizer which is agnostic to input semantics has remained prevalent when training LLMs to additionally process visual information~\citep{beyer2024paligemma, tong2024cambrian1}. 

Multimodal capabilities are the cornerstone of real-world applications including robotics \cite{kim2024openvla}. Recently, Vision-Language-Action models (VLAs) have gained significant traction \cite{brohan2022rt1, brohan2023rt2, kim2024openvla, o2024openx, team2024octo} due to their ability to adapt general knowledge to an imitation-learning setting. VLAs typically take images from cameras in the environment and text, e.g., language instructions, to predict robotic actions, allowing multimodal prompting of agents for robotic manipulation. Despite the success of VLAs, they rely on expensive pre-training on specialized robotics datasets such as \oxe \cite{o2024openx, kim2024openvla, team2024octo} to generalize to new tasks. Unfortunately, training VLAs at this scale requires large amounts of compute resources, unavailable to many research laboratories. 
%We believe that enabling research on large VLAs at scale is of utmost importance to drive further progress on these methods.

We identify the visual processing of current state-of-the-art VLAs \cite{kim2024openvla} as one of the main bottlenecks that significantly increase the compute requirements of training VLAs. This process generally consists of dividing the image into (hundreds of) patches \cite{dosovitskiy2021vit}, processing them through a visual encoder model, and then feeding the outputs as visual tokens to the LLM. As we show in Figure~\ref{fig:training_curves}, this visual tokenization scheme, which so far is the most popularly adopted, can be made fundamentally more efficient without compromising performance on robotics tasks. Our key insight is that task execution given a language instruction requires focus on specific parts of the input observation such as objects of interest, while background information is less relevant and can therefore be discounted. 

% \textbf{Contributions.} 
In this work, we present an Object-Agent-centric Tokenization scheme for VLAs (\ourvla), which adaptively tokenizes important parts of the scene into a reduced set of visual tokens to feed to the VLA. By producing fewer, but more meaningful tokens, \ourvla significantly reduces memory and compute requirements. Furthermore, we designed \ourvla's architecture to be able to leverage the pre-trained knowledge already contained in OpenVLA \cite{kim2024openvla} to maintain easy adaptation capabilities and high fine-tuning performance, which match or even outperform heavier VLA training schemes.

Our contributions can be summarized as follows:
\begin{itemize}[nolistsep]
\item Inspired by object-centric representations, we design a method to condense visual information about objects into a reduced set of tokens;
\item We design a simple approach to identify agent-related visual patches and concatenate them with object-centric tokens to ensure precision of the VLA when predicting actions; 
\item We show that the overall model, \ourvla, is modular and scalable, so that it can be efficiently adapted by reusing existing VLA checkpoints. We assess the performance of our method on the \libero benchmark and demonstrate a 2x speedup in terms of convergence according to the action token accuracy and success rate on tasks;
\item Finally, we tested \ourvla on real-world pick and place tasks and find that it executes more robustly than OpenVLA, translating into a higher success rate.
\end{itemize}
\vspace{-1em}

\section{Related work}
\label{sec:related}
\paragraph{Vision-language-action models.} 
There has been an increasing trend in leveraging generalization capabilities of Vision-Language-Models (VLMs) for robotic control by fine-tuning these VLMs on robotics tasks~\cite{zitkovich2023rt, kim2024openvla, black2410pi0}. RT-2~\cite{zitkovich2023rt} utilizes frontier closed-sourced VLMs by co-fine-tuning them on visual and robotics tasks to achieve a general purpose robotics policy. At the same time, OpenVLA~\cite{kim2024openvla} and $\pi0$~\cite{black2410pi0} embody similar open-source efforts. Preceding these, there have been multiple efforts at using the transformer architecture trained on large robotics dataset to achieve general purpose robotics policies. Gato~\cite{reed2022generalist} presented the first transformer trained on a multitude of language, vision and robotics tasks whereas RT-1~\cite{brohan2022rt1} presented a first concentrated effort at using large transformer architectures trained on a large robotics dataset. Octo~\cite{team2024octo} and JAT~\cite{gallouedec2024jack} present open-source transformer policies designed for efficient fine-tuning on down-stream tasks. 
Consequently, there have been multiple efforts at incorporating spatial awareness~\cite{qu2025spatialvla}, 3D information~\cite{zhen20243d}, generalization to new objects~\cite{zhu2025objectvla}, recurrence and streaming ability~\cite{li2023vision, liu2024robomamba, schmied2024large, haresh2024clevrskills} and embodied chain-of-thought~\cite{Zawalski24-ecot, li2024llara} to VLAs to improve generalization. In our work, we focus on improving the visual representation of VLAs for compute efficiency reasons, while orthogonal works \cite{kim2025fine, reuss2025efficient} have shown significant progress by improving the action prediction component. 

%Visual Question Answering

\paragraph{Efficient VLMs, VLAs.} Concurrent to improving generalization capability of VLAs, a number of works have explored improving efficiency of these models and the VLMs on which they build. Related work TokenLearner~\cite{ryoo2021tokenlearner} introduces a generic learned method to reduce the number of tokens, while other works \cite{performers, sara_rt} increase the efficiency of the attention layers in transformers.  TinyVLA~\cite{wen2025tiny} presents a VLA with a smaller backbone and diffusion-based action heads for faster training and inference speeds. Similarly, MiniVLA~\cite{belkhaleminivla} trains a performant VLA using a 1B parameters models. VLA-Cache~\cite{xu2025vla} proposes a caching mechanism that adaptively identifies visual tokens with minimal changes and re-uses computation for ``unchanged'' visual tokens for efficient processing. Deer-VLA~\cite{yue2024deer} on the other hand, proposes dynamic inference that allows automatically adjusting the size of the model leading to performance improvements in inference. Previously, improving VLA tokenization has been studied by \cite{pertsch2025fast}, which we consider orthogonal to our work as we aim to improve the visual backbone as opposed to the action tokenization. 

\paragraph{Object-centric representations for robotics.}

Object-centric representation learning focuses on modeling individual objects within a scene to improve understanding of complex environments. Aside from reinforcement learning \cite{ferraro2023focus, sancaktar2022structuredWM, yoon2023ocrl, haramati2024entity}, it has gained traction in robotics and imitation learning: A3VLM~\cite{huang2024a3vlm} leverages an object-centric approach to train robot agnostic VLAs, while DexGraspVLA~\cite{zhong2025dexgraspvla} utilizes SAM~\citep{ravi2024sam} to additionally feed object-mask features into a VLA. Similarly, Sam2Act ~\cite{fang2025sam2act} uses SAM as an image encoder to learn multi-view policies, FOCUS~\cite{ferraro2023focus} obtains mask supervision from SAM and POCR~\cite{shi2024PT-OC-IL} proposes a general object-centric framework for pre-trained robotic embeddings. Concurrent with our work, ControlManip~\cite{li2025controlmanip} conditions a pre-trained VLA on an object-centric representation. For our work, we employ FT-Dinosaur~\cite{didolkar2024ftdino}, which employs an unsupervised object-centric approach to adapt vision encoders for object discovery.

\section{Method}
\label{sec:method}

VLAs learn a distribution $p(\mathbf{a} | \mathbf{o}, \ell)$ over actions $\mathbf{a}$, given an observation $\mathbf{o}$ and a task description $\ell$. In our case, we assume that $\mathbf{o} \in \mathbb{R}^{C\times W \times H}$  consists of a single camera image. Observations are encoded using a visual encoder $\text{VisEnc}$, such as \dinosiglip~\cite{oquab2023dinov2, zhai2023sigmoid}, resulting in a set of visual tokens $\mathbf{v}_{1\dots K} = \text{VisEnc}(\mathbf{o})$, whereas language instructions are encoded using the tokenizer of the underlying LLM, which we denote as $\mathbf{l}_{1\dots J}$. Overall, we have \(p(\mathbf{a} | \mathbf{o}, \ell) = \text{LLM}(\mathbf{a}|\mathbf{l}_1,\dots,\mathbf{l}_J,\mathbf{v}_1,\dots, \mathbf{v}_K)\), where the distribution over actions is chosen to be discrete by using a binning scheme \cite{kim2024openvla}. %  with 256 bins

The most common visual tokenization strategy consists of dividing the observations into patches, processing them with a visual encoder, and feeding them to the LLM. For example, a $224\times224$ pixel image can be divided into 256 patches of size $14\times14$ pixels. This would result in the LLM processing $K=256$ visual tokens, an amount of tokens that is generally one order of magnitude larger than the amount of tokens processed for a language instruction. As a consequence, the amount of visual tokens is one of the main bottlenecks when training VLAs. The aim of our work is to greatly reduce the number \(K\) of visual tokens with no loss of performance. 

% To do so, we present two architectures of varying complexity.

\subsection{Object-centric tokens}
\label{subsec:oc_tokens}

In order to understand the scene, the agent needs to be able to look at all parts of the observed images. While feeding embedded image patches to the VLA's LLM is a safe design choice, it is inefficient from an information perspective. Several patches in an image often contain information that is irrelevant for the agent, e.g., background information. Furthermore, many patches in an image can represent the same entities, e.g., an object spanning across multiple patches. Such information could be efficiently compressed into a smaller amounts of tokens, which we call \textit{object-centric tokens}.

Object-centric models that perform semantic segmentation \cite{ravi2024sam, didolkar2024ftdino} allow to group pixels in an image, based on their common semantical meaning. Each group of pixels is generally assigned an index and can be referred to as an "object mask". Given the observation $\mathbf{o}$, we process the image with an object extractor, to obtain $N$ segmentation masks, i.e. $\mathbf{m}_{1...N} = \text{ObjEnc}(\mathbf{o})$, where $N$ can be a fixed or a variable number.

In order to obtain \textit{object-centric tokens} for our VLA, we perform the following operations:
\begin{enumerate}[nolistsep]
    \item we extract \textit{visual tokens} for all patches in an image, using the visual encoder;
    \item we obtain the \textit{object masks} using an object-centric model at the same output resolution as the patches, and gather the visual tokens based on their corresponding masks;  
    \item we perform a \textit{pooling} operation to compress the information of the tokens that belong to the same object mask.   
\end{enumerate}

Formally, we compute object-centric tokens as
\begin{align*}
\mathbf{t}_j &= \text{pool}( \left\{\mathbf{t}_n^k \mid {k\in\{1\dots K\},}\ n\in\{1\dots N\} \right\}), \\ \mathbf{t}_n^k &= \mathbf{m}_n^k\odot \mathbf{v}_k ,
\end{align*}
where \(\mathbf{m}_n^k\) is the mask corresponding to object $n$ indexed by the patch coordinates of \(\mathbf{v}_k\). This procedure allows us to reduce the number of visual tokens processed by 1 or 2 orders of magnitude, depending on the masking strategy adopted. In our experiments, we adopt FT-Dinosaur \cite{didolkar2024ftdino}, an unsupervised object-centric model. FT-Dinosaur has two features that make it a good fit for our approach: (i) it can be flexibly fine-tuned on new datasets in an unsupervised fashion, and (ii) it allows to define a fixed number of masks to be extracted from an image. For the pooling part, we adopt a simple average pooling and ablate this choice in Section \ref{sec:ablations}. For our main experiments, we use only 7 object masks, consequently obtaining 7 object-centric tokens.
\subsection{Agent-centric tokens}
\label{subsec:ac_tokens}
\begin{figure}[t]
    \centering
    \includegraphics[width=0.95\linewidth]{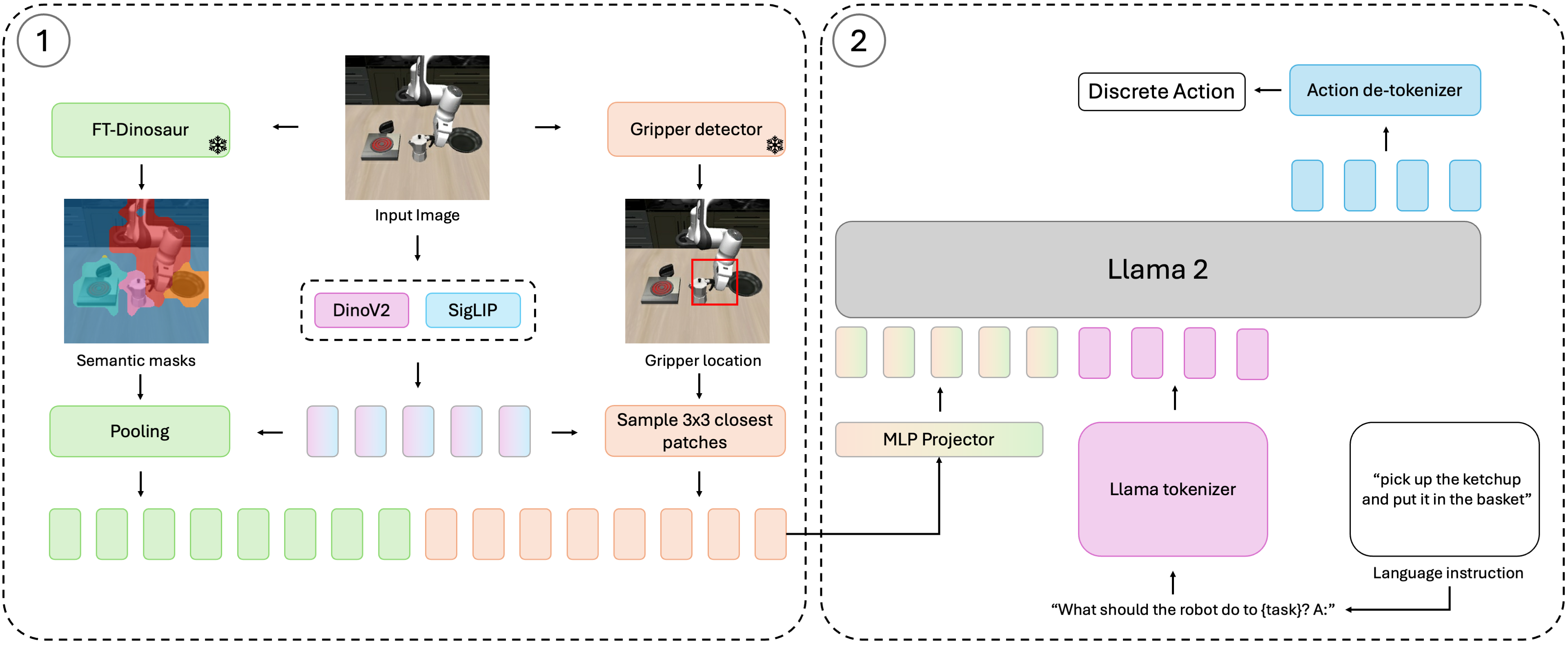}
    \caption{\ourvla introduces a visual tokenization process, which extracts object-centric and agent-centric tokens. These tokens are then fed to the LLM for action prediction.}
    \label{fig:oat_vla_architecture}
\end{figure}

While object-centric tokens summarize \emph{what} is present and \emph{where}, they inherently discard high-frequency details. For fine-grained manipulation, it is crucial that the agent possesses accurate information about the end effectors and their interactions with objects.
During interaction with objects, the objects and the end effectors' information may be aggregated into different object-centric tokens. One of the consequences is that information about the interaction, e.g., touching an object, may be incidentally discarded. To avoid such possible scenarios, and always provide the agent with high-resolution information about the end-effectors, we introduce \textit{agent-centric tokens}.

In a single-arm robotic manipulation system, the end effector is generally a gripper attached to the arm. Identifying the gripper in the camera observations enables to locate where the agent's actions are being executed. In systems with camera calibration, the gripper location can be obtained from the robot pose. However, such calibration is often unavailable, e.g., it is missing for several datasets in the \oxe dataset~\cite{o2024openx}.  
To obtain a generic solution, we train a gripper detector model. At test time, it returns a single 2-D keypoint, locating the gripper in the visual observation. %, i.e. $(g_x, g_y) = AgDet($.

In order to obtain \textit{agent-centric tokens} for our VLA, we perform the following operations:
\begin{enumerate}[nolistsep]
    \item (shared step with object-centric tokens) we extract \textit{visual tokens} for all patches in an image, using the visual encoder;
    \item we detect the \textit{gripper location} in the camera observation, using the gripper detector;
    \item we obtain a grid of \textit{patches around the agent}, by selecting the patches around the gripper.   
\end{enumerate}
This procedure allows us to select an arbitrary number of visual tokens around the agent's end effector location, which ensures the agent sees the effects of its actions at all times.
In our experiments, we adopt a gripper detector based on a lightweight ResNet-based \cite{he2015resnet} Faster R-CNN~\cite{ren2015faster} architecture. The gripper training is supervised with a few thousand manually annotated frames from \oxe\ plus automatically extracted gripper locations for the \libero dataset. For the agent-centric tokens, we adopt a $3\times3$ patches grid, which results in a set of 9 agent-centric tokens. In the very rare cases where no gripper is detected, the agent-centric tokens take the value of all patch features over the entire image.

\subsection{\ourvla: Object-Action-centric Tokenization for VLAs}

Our method, \ourvla, aims to provide a more efficient VLA training strategy, by employing a reduced number of object-centric and agent-centric tokens. In \ourvla, object-centric tokens enable semantically-organized perception of the scene, while agent-centric tokens enable precise manipulation. \ourvla's visual tokenization process is illustrated in Figure \ref{fig:oat_vla_architecture}.

The visual tokenization is the main innovation of our approach. It consists of the object-centric tokens extraction and the agent-centric tokens extraction procedures, as explained in this section.
The resulting tokens are passed through an MLP projector and fed to the (Llama 2) LLM backbone, following the OpenVLA's architecture ~\citep{kim2024openvla}. Crucially, \ourvla does not disruptively change the information contained in the tokens fed to the LLM, as it only performs operations of aggregation and selection of the tokens from the pre-trained visual encoder. Thus, as we also show in Section \ref{sec:result}, the agent is still able to leverage the pre-trained knowledge coming from the base VLM/VLA. 

For 224x224 images, the number of visual tokens used by \ourvla is 16 in total: 7 object-centric tokens, and 9 agent-centric tokens. For the same image resolution, the original OpenVLA's visual backbones extracts 256 visual tokens. Thus, \ourvla uses \textbf{\textcolor{black}{93.75\% less visual tokens}}. This reduction allows to increase the batch size significantly during training and, as we observe in Figure \ref{fig:training_curves}, it allows to train our VLAs to achieve slightly higher performance than OpenVLA, and  \textbf{\textcolor{black}{more than $2\times$ faster}} (both in terms of training steps and actual time).

\section{Experimental results}
\label{sec:result}

In Figure \ref{fig:training_curves}, we show that \ourvla can be trained significantly faster than its "original" counterpart, the OpenVLA model. The goal of this section is to demonstrate in detail that \ourvla's performance is consistently in line or superior to OpenVLA, across different training settings.

In terms of environments, we adopt two settings: the \libero environment, with its four task suites, and a real-world environment, using a \xarm to rearrange objects on a tabletop. 

For OpenVLA experiments below, we start training our models from the the OpenVLA checkpoint \cite{openvlacheckpoint} pre-trained on the \oxe dataset (which excludes the \libero dataset). For \ourvla experiments, we start from the OpenVLA checkpoint that has been fined-tuned using the \ourvla architecture on a subset \oxe for 200K steps, see Appendix~\ref{sec:appendix_pre_training} for details. We leave training \ourvla from a stock VLM checkpoint such as Prismatic VLMs \cite{karamcheti2024prismaticvlm} or PaliGemma \cite{beyer2024paligemmaversatile3bvlm} for future work.

\subsection{Full Fine-tuning}
\label{subsec:full_fine_tuning}

% This enables us to compare to the results in Appendix~E of OpenVLA~\cite{kim2024openvla}. 

We want to show that \ourvla enables a more efficient fine-tuning of the all the weights of the VLA model on new datasets. For this purpose, starting from the corresponding base VLA checkpoints, we fine-tune both \ourvla and OpenVLA on the full \libero dataset (\textsc{Spatial}, \textsc{Object}, \textsc{Goal}, \textsc{10} and \textsc{90}). We use a single action space normalization for all five subsets. 

Full fine-tuning of \ourvla is performed using a batch size of $8 \times 64=512$ which we measured to process 320 examples/second on average on an 8xH100 node.
Full fine-tuning of OpenVLA is performed using a batch size of $8 \times 32=256$, processing 157 examples/second on average on an 8xH100 node. The larger batch-size of \ourvla is in inherent advantage of the method, because fewer visual tokens result in lower GPU memory requirements per sample. Any other training settings are chosen to remain the same as in OpenVLA. 

The models are evaluated on the four \libero task suites: \textsc{Spatial} tests the agent’s understanding of spatial relationships, \textsc{Object} tests the agent’s understanding of object types, \textsc{Goal} tests the agent’s knowledge of different task-oriented behaviors and \textsc{Long}, also referred to as \textsc{10} consists of a variety of long-horizon tasks with diverse object interactions and versatile motor skills.  For each task suite, we run 100 evaluations.

\begin{figure}[t!]
    \centering
    \includegraphics[width=\linewidth]{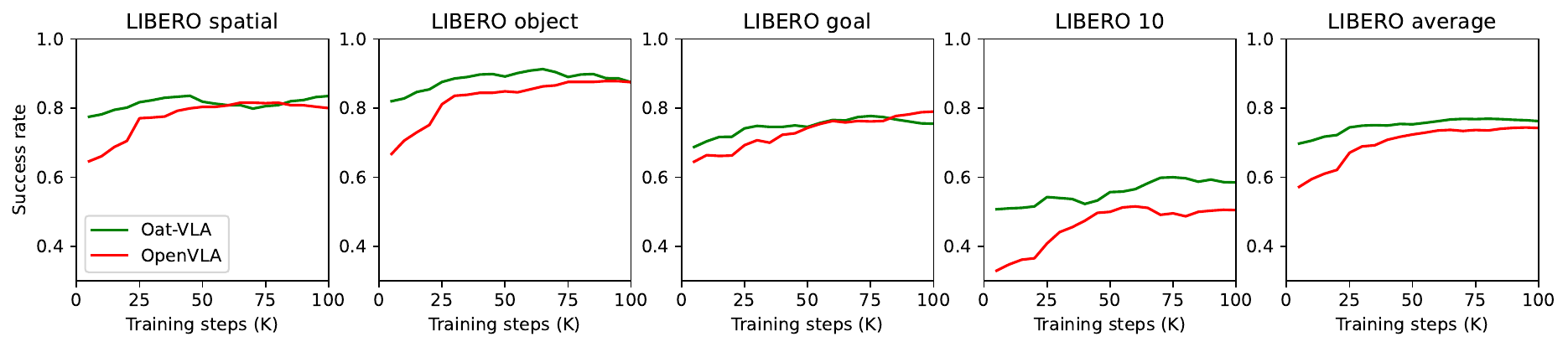}
    \caption{Evaluations on \libero, every 5K training (full fine-tuning) steps. The plots are mean filtered using one evaluation before and after.}
    \label{fig:full_finetuning_libero_evaluation}
\end{figure}

We used the full fine-tuning to study how the action token accuracy evolves over time. This is represented  in Figure \ref{fig:training_curves}.
% and \libero success rates  \ourvla and OpenVLA. 
While training the models, we also evaluate checkpoints every 5000 training steps, to benchmark the agent's performance in terms of training efficiency. 
We report the performance of the two models over training steps in Figure \ref{fig:full_finetuning_libero_evaluation}. Overall, \ourvla performs slightly better than OpenVLA, with a more significant performance advantage in the challenging \libero 10 suite. Importantly, \ourvla obtains higher success rates more than $2\times$ faster, thanks to the higher batch size and improved visual encoding, which simplifies training by discarding or compressing less relevant information.

\begin{table}[b]
\centering
\caption{LoRA fine-tuning success rate comparison on \libero. The numbers for OpenVLA, Octo and Diffusion Policy are taken from OpenVLA~\cite{kim2024openvla}.}
\small{
%\begin{tabular}{rcccccc}
%\toprule
%\libero task & \ourvla & OpenVLA & Octo & Diffusion Policy \\ 
% \cmidrule(lr){2-3}
 % & 30K steps & best &  &  &  \\ 
%\midrule
%Spatial & \textbf{87.3} \tiny{$\pm0.7$}\% & 84.7 \tiny{$\pm0.9$}\% & 78.9 \tiny{$\pm1.0$}\% & %78.3 \tiny{$\pm1.1$}\% \\ 
%Object &  89.1 \tiny{$\pm0.2$}\% & 88.4 \tiny{$\pm0.8$}\% & 85.7 \tiny{$\pm0.9$}\% & \textbf{92.5} \tiny{$\pm0.7$}\% \\ 
%Goal &  82.1 \tiny{$\pm0.8$}\% & 79.2 \tiny{$\pm1.0$}\% & \textbf{84.6} \tiny{$\pm0.9$}\% & 68.3 \tiny{$\pm1.2$}\% \\ 
%10 &  \textbf{55.9} \tiny{$\pm1.8$}\% & 53.7 \tiny{$\pm1.3$}\% & 51.1 \tiny{$\pm1.3$}\% & 50.5 \tiny{$\pm1.3$}\% \\ 
%\midrule
%average & \textbf{78.6} \tiny{$\pm0.5$}\% & 76.5 \tiny{$\pm0.6$}\% & 75.1 \tiny{$\pm0.6$}\% & 72.4 \tiny{$\pm0.7$}\% \\ 
%\bottomrule
%\end{tabular}

\begin{tabular}{rccccc}
\toprule
 & Spatial & Object & Goal & 10 & Average \\
\midrule
\ourvla & \textbf{87.3} \tiny{$\pm0.7$}\% & 89.1 \tiny{$\pm0.2$}\% & 82.1 \tiny{$\pm0.8$}\% & \textbf{55.9} \tiny{$\pm1.8$}\% & \textbf{78.6} \tiny{$\pm0.5$}\% \\
OpenVLA & 84.7 \tiny{$\pm0.9$}\% & 88.4 \tiny{$\pm0.8$}\% & 79.2 \tiny{$\pm1.0$}\% & 53.7 \tiny{$\pm1.3$}\% & 76.5 \tiny{$\pm0.6$}\% \\
Octo & 78.9 \tiny{$\pm1.0$}\% & 85.7 \tiny{$\pm0.9$}\% & \textbf{84.6} \tiny{$\pm0.9$}\% & 51.1 \tiny{$\pm1.3$}\% & 75.1 \tiny{$\pm0.6$}\% \\
Diffusion Policy & 78.3 \tiny{$\pm1.1$}\% & \textbf{92.5} \tiny{$\pm0.7$}\% & 68.3 \tiny{$\pm1.2$}\% & 50.5 \tiny{$\pm1.3$}\% & 72.4 \tiny{$\pm0.7$}\% \\
\bottomrule
\end{tabular}
\label{tab:ablation_libero_task_transposed}

}
\label{tab:libero_lora_performance_v2}
\end{table}

\subsection{LoRA Fine-tuning}
\label{subsec:lora_fine_tuning}

It is desirable that a VLA can be fine-tuned using parameter-efficient techniques, such as LoRA \cite{hu2022lora}. Thus, we compare \ourvla with OpenVLA, when it comes to LoRA fine-tuning of the weights. These results allow for a direct comparison with the LoRA results in \cite{kim2024openvla}.

LoRA fine-tuning is performed using a batch size of $8\times48=384$ for \ourvla  and $8\times16=128$ on an 8xH100 node. Despite the lower number of learnable parameters, the batch-size is smaller than during full fine-tuning because FSDP~\cite{zhao2023pytorchfsdpexperiencesscaling} is not enabled during LoRA fine-tuning in OpenVLA. Other training settings are the same as in OpenVLA. At the given settings, \ourvla processes 384 examples/second on an H100 node, while OpenVLA processes 197 examples/second. 

\begin{figure}[t]
    \centering
    \caption{\textbf{Top.} The setup for some of the real-world tasks. (a) banana in green bowl (b) red cube in brown bag (c) zucchini in front of green cube (d) tomato left of lettuce. \textbf{Bottom.} The table reports the success rates on the real-world tasks and number of successful trials. }
    \centering
    \begin{subfigure}[t]{0.18\textwidth}
    \centering
    \includegraphics[width=\linewidth]{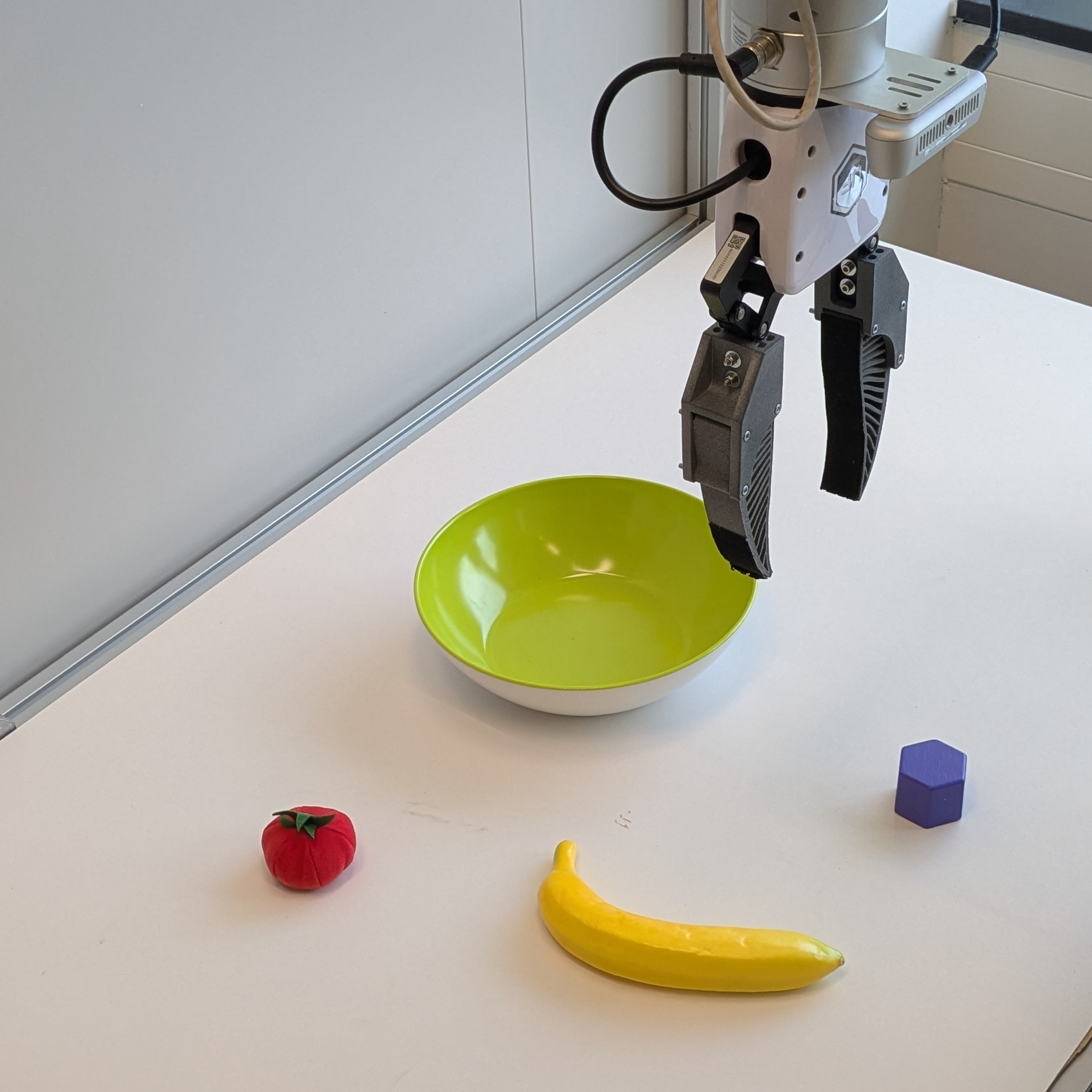}    
    \caption{}
    \end{subfigure}
    \hfill
    \begin{subfigure}[t]{0.18\textwidth}
    \centering
    \includegraphics[width=\linewidth]{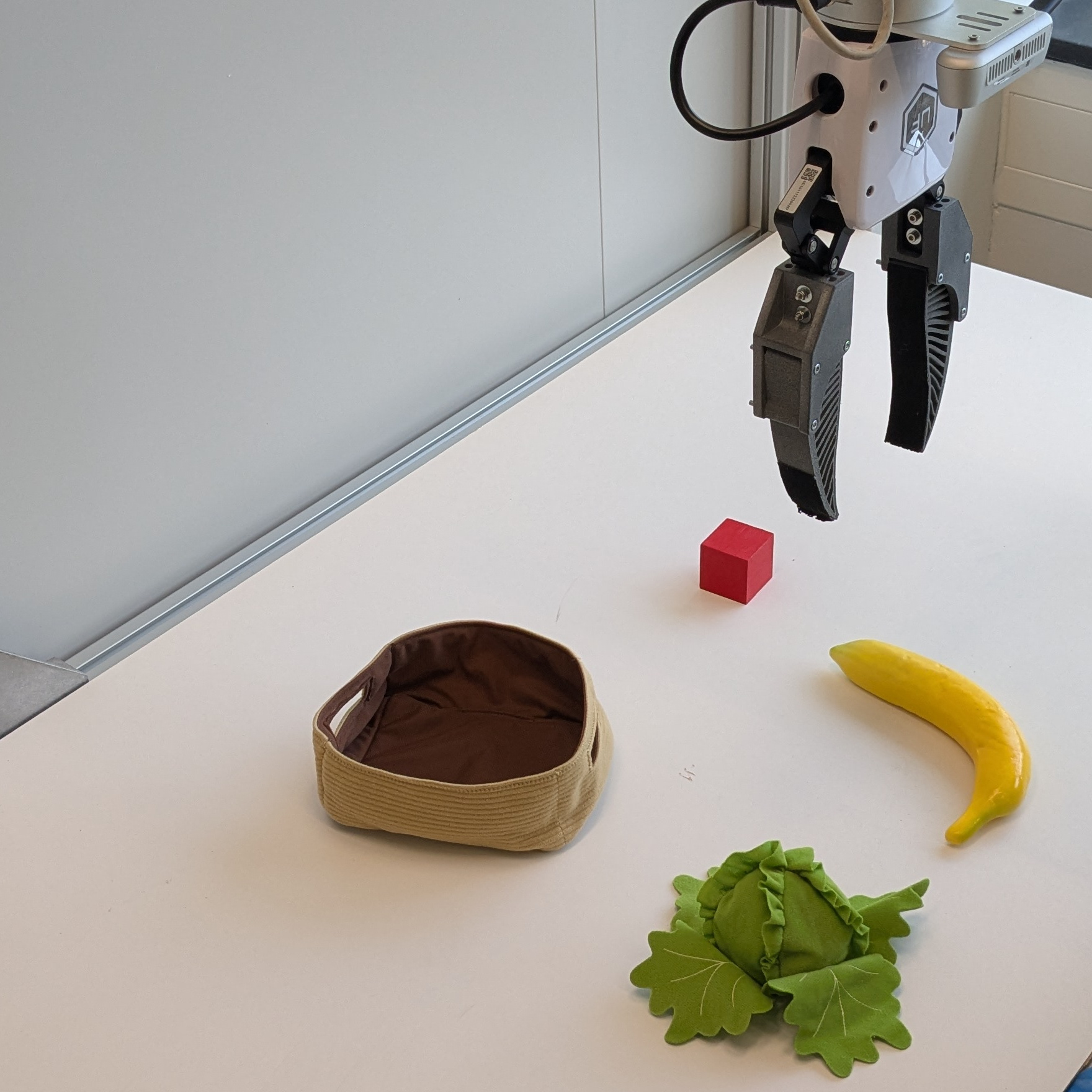}    
    \caption{}
    \end{subfigure}
    \hfill
    \begin{subfigure}[t]{0.18\textwidth}
    \centering
    \includegraphics[width=\linewidth]{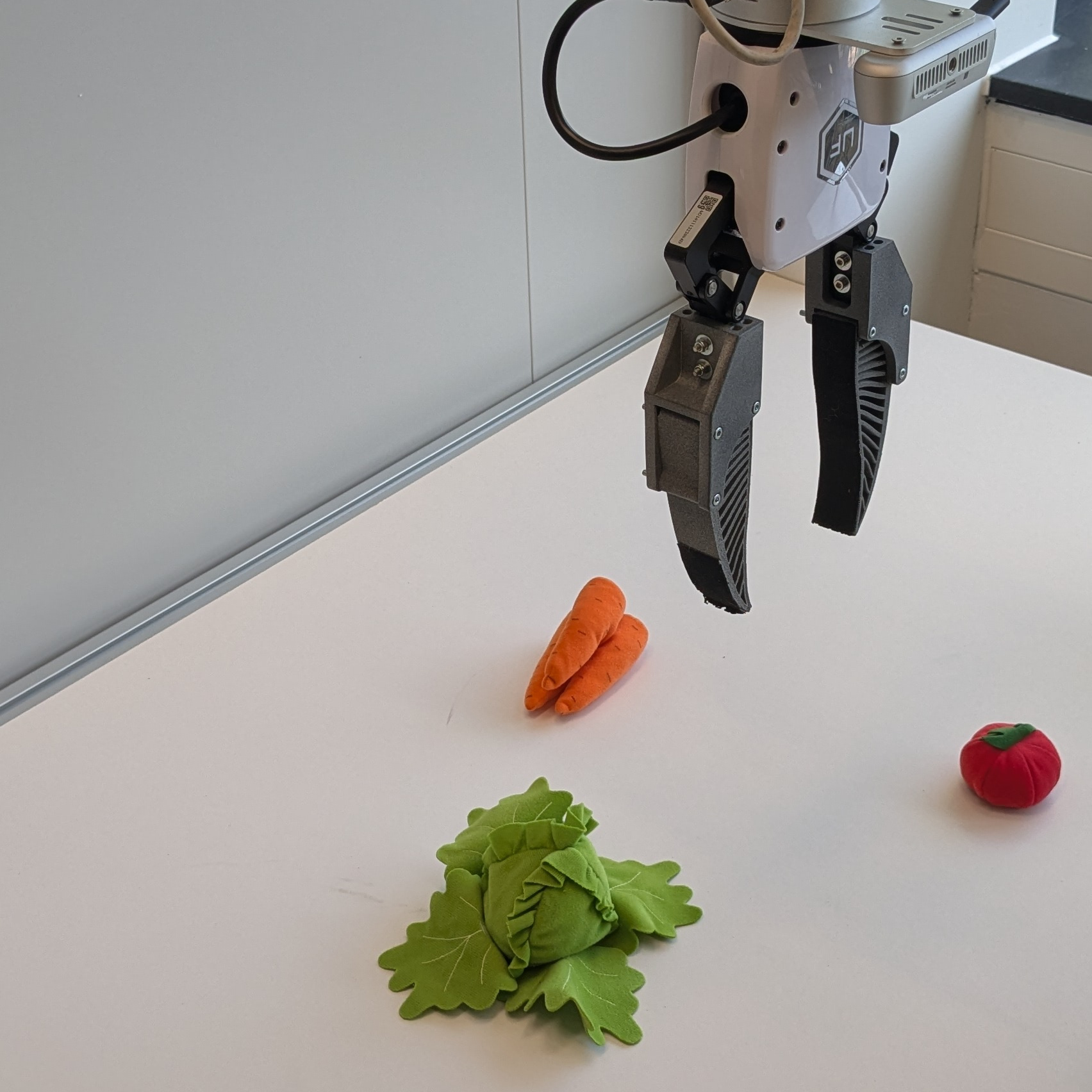} 
    \caption{}
    \end{subfigure}
    \hfill
    \begin{subfigure}[t]{0.18\textwidth}    
    \centering
    \includegraphics[width=\linewidth]{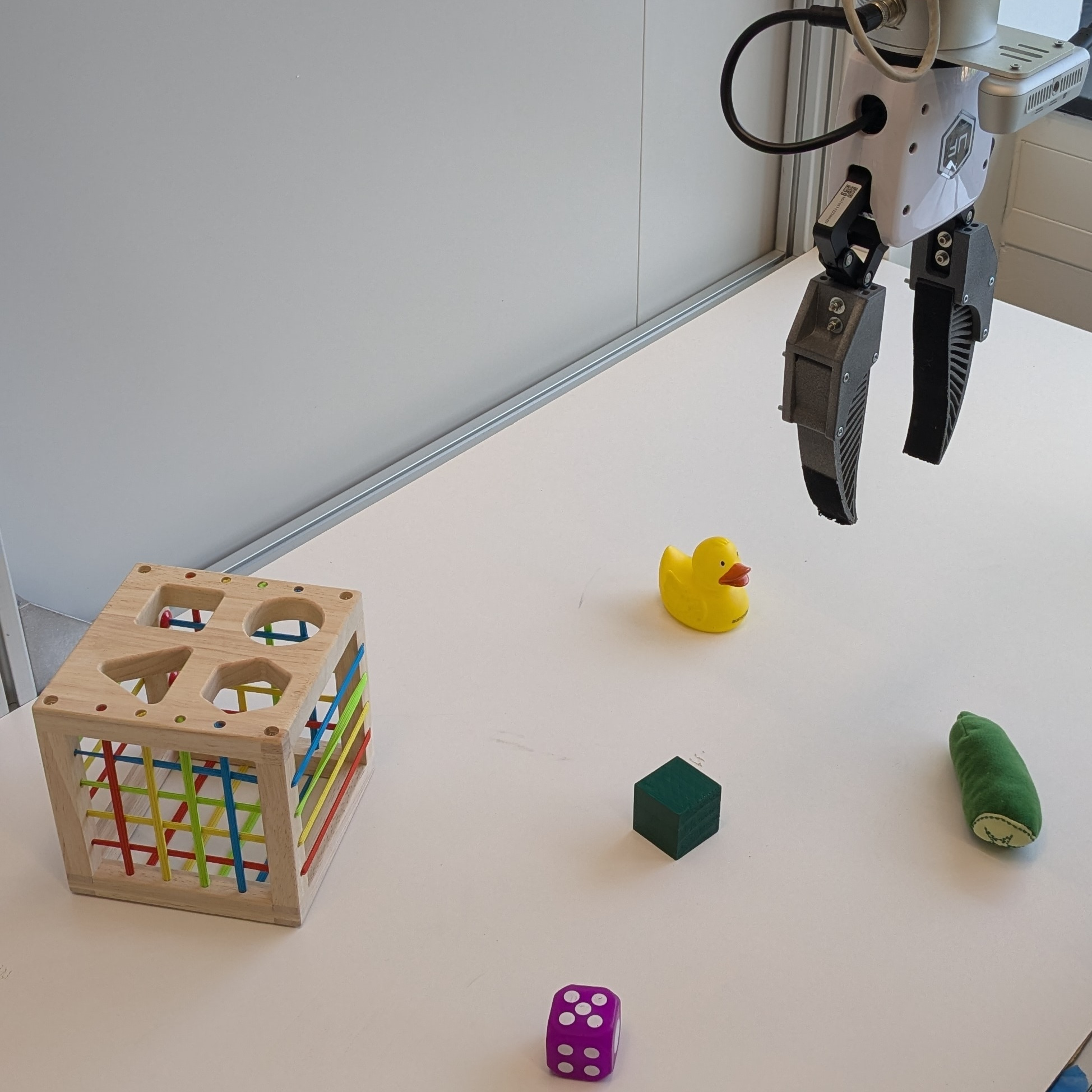}    
    \caption{}
    \end{subfigure}
    \small{
\begin{tabular}{lcc}
\toprule
 & \textbf{OpenVLA} & \textbf{\ourvla} \\
\midrule
\textbf{In-distribution tasks} & \textbf{52}\% (13/25) & \textbf{72}\% (18/25) \\ \midrule
Place the banana in the green bowl & 70\% (7/10) & 70\% (7/10) \\
Place the red cube in the brown bag & 50\% (3/6) & 33\% (2/6) \\
Place the tomato left of the lettuce & 60\% (3/5) & 100\% (5/5) \\
Place the zucchini in front of the green cube & 0\% (0/4) & 100\% (4/4) \\
\midrule
\textbf{Out-of-distribution tasks} & \textbf{29}\% (7/24) & \textbf{46}\% (11/24) \\ \midrule
Place the rubber duck in the green bowl & 40\% (4/10) & 30\% (3/10) \\
Place the mushroom in the brown bag & 0\% (0/6) & 50\% (3/6) \\
Place the purple die left of the lettuce & 0\% (0/4) & 50\% (2/4) \\
Place the zucchini in front of the red hexagon & 75\% (3/4) & 75\% (3/4) \\
\midrule
\textbf{Overall} & \textbf{41}\% (20/49) & \textbf{59}\% (29/49)\\
\bottomrule
\end{tabular}
}
    \label{fig:real_world}
\end{figure}

\paragraph{\libero benchmark.}

Following OpenVLA~\cite{kim2024openvla}, evaluations are performed using three seeds with 500 roll-outs per task per seed and we report average success rates and standard deviation. Our results are in Table~\ref{tab:libero_lora_performance_v2}, where we also compare to the results of Octo~\cite{team2024octo} and Diffusion Policy~\cite{chi2023diffusion}. Similarly to the full fine-tuning results, the performance of \ourvla is slightly higher than OpenVLA, in all the task suites, increasing the overall advantage over the other methods. We note that \ourvla achieves good performance at 30K training steps (see Appendix~\ref{sec:appendix_full_fine_tuning}), while the released OpenVLA checkpoints~\cite{openvlacheckpoint} are around 60K steps. Both \ourvla and OpenVLA have lower performance than Octo in Goal and than Diffusion Policy in Object. This may be due to the diffusion process adopted in these methods, which provides more accurate action prediction.

\paragraph{Real-world benchmark.} We collect a dataset of 320 trajectories on a robotic setup using a \xarm operating on a white tabletop. The setup is observed by the agent through the lenses of a RealSense D435 camera positioned in a corner of the table, using only the RGB image.

We evaluate on two sets of tasks: an in-distribution set and an out-of-distribution set. For the in-distribution tasks, at least 10 demos of such tasks are present in the dataset. For out-of-distribution tasks, we take the in-distribution tasks and swap some of the objects, e.g., the object to pick or place next to, so that the agent never saw such tasks in the dataset. For all the evaluations, we consistently re-position the objects with slight variations, e.g., different orientations or positions of an object. Further details are provided in Appendix~\ref{sec:appendix_real_world}.

Visualizations of some of tasks and the results are shown in Figure \ref{fig:real_world}. \ourvla reports a consistent advantage in performance over OpenVLA both on in-distribution and out-of-distribution tasks. From qualitative assessment of the robot executions, we observed that \ourvla tends to be more precise at picking and placing objects. Instead, OpenVLA would sometimes grasp in the air above the target object or place the object on the wrong side/on top of the target location.

\subsection{Ablations}
\label{sec:ablations}

With our ablation study, we analyze different ways of condensing the information in the visual tokens and verify some of our design choices.
We adopt the full fine-tuning setup on the \libero benchmark and train all models for 20K training steps. The ablations we analyze are as follows:
\begin{itemize}[nolistsep]
    \item \textbf{Single image token}: condensing information from all visual tokens into a single token, using attention pooling \cite{radford2021clip};
    \item \textbf{Object-centric tokens only}: extracting the object tokens, using the method described in Section \ref{subsec:oc_tokens}. Using attention pooling to condense information;
    \item \textbf{\ourvla (attention pooling)}: using both object-centric tokens and agent-centric tokens, where the object-centric pooling is an attention pooling layer;
    \item \textbf{\ourvla (average pooling)}: object-centric tokens and agent-centric tokens, where object-centric pooling is an average pooling layer (default method for all other experiments).
\end{itemize}

The results are presented in Table \ref{tab:ablation_visual_tokens}. We observe that condensing information into multiple tokens for the objects, rather than into a single token, improves performance slightly, especially in \libero Spatial. The agent-centric tokens have a major positive effect on performance and, as we hypothesized, they are crucial to not incur any loss in performance. Finally, we found average pooling to work better than attention pooling, possibly due to its simplicity, i.e. no additional layers. We believe that other pooling strategies could be used to better condense the object token information, but we leave such direction for future work.

\begin{table}[h!]
\centering
\caption{Ablation experiments for design choices in \ourvla on \libero}
\small{
\begin{tabular}{rccccc}
\toprule
& Spatial & Object & Goal & 10 & Average \\
\midrule
Single image token & 64.0\% & 74.0\% & 71.0\% & 31.0\% & 60.0\% \\
Object-centric tokens only & 75.0\% & 74.0\% & 67.0\% & 29.0\% & 61.3\% \\
% One Image Token + Agent-centric & 80.0 & 88.0 & 69.0 & 44.0 & 70.3 \\
\ourvla (attention pool) & 81.0\% & 86.0\% & 73.0\% & 41.0\% & 70.3\% \\
\ourvla (average pool) & \textbf{86.5\%} & \textbf{89.1\%} & \textbf{80.3\%} & \textbf{52.5\%} & \textbf{77.1\%} \\
\bottomrule
\end{tabular}
}
\label{tab:ablation_visual_tokens}
\end{table}

\section{Discussion and conclusion}
\label{sec:discussion}

Training VLAs at scale is an important problem as we progress towards multimodal robotics systems that handle both visual and language inputs. However, training VLAs comes with several challenges, particularly in terms of large compute resource requirements and long training times. With \ourvla, we propose a way to drastically reduce the training cost and time of VLAs by reworking the visual tokenization process using object-centric and agent-centric tokens.  

Our approach implicitly introduces some inductive bias into the system. For the agent-centric tokens, we used the end-effector as a straightforward proxy for "the location where the agent interacts with the world". This should be generalized to allow multiple manipulators and non-prehensile manipulation with other parts of the robot. For the object-centric tokens, we employ object-centric models that perform semantic segmentation. Future work should aim to find object-agent-centric-like inductive biases that arise more naturally from the data.

In conclusion, we hope to inspire more work in the area of efficient visual tokenization for VLAs and other end-to-end trained robot policies. Additionally, we hope that our work will enable more research labs to work on the training of foundational VLA models.

\section{Limitations}
\label{sec:limitations}

We have only tested \ourvla on a single-armed robot for pick and place tasks. It remains to be seen how \ourvla performs in a bi-manual setting, and on more complex tasks such as folding clothes.

Our prototype consists of three distinct models: FT-Dinosaur for the object-centric masks, Faster R-CNN for the gripper detector and OpenVLA as the VLA. It makes sense to let all parts share the same visual backbone. We do not expect any changes in success rates, but training time should improve in the order of 5\% to 10\%.

As shown in Appendix~\ref{sec:appendix_inference_time}, using A100 or RTX A5000 GPUs, \ourvla offers no advantage in batch-of-1 inference speed, because in our analysis the inference time is dominated by loading the LLM weights from GPU RAM to GPU cache. This might change with different hardware architectures, or different action heads where the batch-size during roll-out increases due to parallel inference, such as for example in OpenVLA-OFT~\cite{kim2025fine}. 

\bibliography{biblio}  % .bib

\clearpage
\newpage
\setcounter{section}{7}
\section*{Supplementary Material}

\subsection{Gripper detector}

As mentioned in Section~\ref{subsec:ac_tokens}, \ourvla requires a gripper detector to identify the agent-centric tokens. We manually annotated the gripper location for 2000 images for three \oxe subsets (Bridge,  Fractal, FMB), and another 2000 annotated images from our own real-world dataset (see Section~\ref{subsec:lora_fine_tuning}) and added in 50K \libero images, with known gripper locations from the simulator. The annotations indicate the end-effector position, i.e. the point midway between the two gripper-fingers. We then fine-tune a ResNet R-CNN~\cite{ren2015faster} to detect a fictional $64\times32$ pixel bounding box around the end-effector position. The IoU of the resulting detector is $> 80\%$ on the holdout set.

\subsection{Pre-training Oat-VLA on Open X-Embodiment details}
\label{sec:appendix_pre_training}

The fine-tuning experiments in Section~\ref{sec:result} are initialized from 
an \ourvla checkpoint that has been pre-trained on a subset of \oxe as follows: starting from the weights of the pre-trained OpenVLA checkpoint~\cite{openvlacheckpoint} (LLama-2 7B, DINOv2 ViT-L/14 and SigLIP ViT-So400M/14, resolution $224 \times 224$), we fine-tune \ourvla to around 95\% action token accuracy on the Bridge+FMB+Fractal subset of \oxe for 235K steps.

\begin{figure}[h]
    \centering
    \includegraphics[width=8cm]{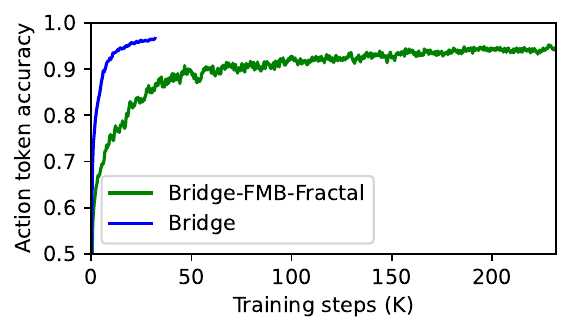}
    \caption{Training action token accuracy while fine-tuning \ourvla on \oxe subsets, starting from the OpenVLA checkpoint. The plots are mean filtered using 250 training steps before and after.}
    \label{fig:bridge_action_token_accuracy}
\end{figure}

Figure~\ref{fig:bridge_action_token_accuracy} shows the training action token accuracy as a function of training steps. Additionally, it shows a second experiment where we fine-tune on only the Bridge dataset. We note that \ourvla gets to high accuracy in 30K steps for Bridge, but the combined Bridge+FMB+Fractal dataset takes longer to reach 95\% action token accuracy mark, i.e. about 10\% of the compute budget used to train the original OpenVLA checkpoint~\cite{kim2024openvla}.

\subsection{Detailed LoRA fine-tuning results}

Table~\ref{tab:libero_lora_performance_v2} in Section~\ref{subsec:lora_fine_tuning} reports the highest success rate of \ourvla across several checkpoints (i.e. 20K, 30K, 40K, 50K training steps). This reporting style follows OpenVLA~\cite{openvlacheckpoint} which hand-picked the checkpoints for \libero Spatial (50K training steps), Object (50K training steps), Goal (60K training steps), and 10 (80K training steps). For transparency, Table~\ref{tab:libero_lora_performance_full} shows the success rate of \ourvla for each individual checkpoint. We note that \ourvla reaches very good performance across all \libero task suites at 30K training steps. In Table~\ref{tab:libero_lora_performance_full_32x8}, we additionally show results for LoRA fine-tuning with a batch size of $8\times32$.

\begin{table}[h]
\centering
\caption{\textbf{LoRA fine-tuning (8$\times$48 batch size)} success rates evaluated for different training steps.
500 evaluations per \libero task suite, 3 seeds. OpenVLA results copied from \cite{kim2024openvla}.}
\begin{tabular}{rcccccc}
\toprule
 & \multicolumn{4}{c}{\ourvla} & \multicolumn{1}{c}{OpenVLA} \\ 
\cmidrule(lr){2-5} \cmidrule(lr){6-6}
\libero task suite & 20K steps & 30K steps & 40K steps & 50K steps &  \\ 
\midrule
spatial & 86.5 \tiny{$\pm1.4$}\% & \textbf{87.3} \tiny{$\pm0.7$}\% & 85.8 \tiny{$\pm0.9$}\% & 85.3 \tiny{$\pm0.6$}\% & 84.7 \tiny{$\pm0.9$}\% \\ 
object & \textbf{89.1} \tiny{$\pm0.2$}\% & 88.2 \tiny{$\pm0.6$}\% & 87.2 \tiny{$\pm2.0$}\% & 88.1 \tiny{$\pm0.7$}\% & 88.4 \tiny{$\pm0.8$}\% \\ 
goal & 80.3 \tiny{$\pm1.1$}\% & 81.7 \tiny{$\pm0.9$}\% & 80.3 \tiny{$\pm0.4$}\% & \textbf{82.1} \tiny{$\pm0.8$}\% & 79.2 \tiny{$\pm1.0$}\% \\ 
10 & 52.5 \tiny{$\pm0.5$}\% & \textbf{55.9} \tiny{$\pm1.8$}\% & 52.7 \tiny{$\pm1.8$}\% & 54.7 \tiny{$\pm1.0$}\% & 53.7 \tiny{$\pm1.3$}\% \\ 
\midrule
average & 77.1 \tiny{$\pm0.2$}\% & \textbf{78.3} \tiny{$\pm0.3$}\% & 76.5 \tiny{$\pm0.9$}\% & 77.6 \tiny{$\pm0.3$}\% & 76.5 \tiny{$\pm0.6$}\% \\ 
\bottomrule
\end{tabular}
\label{tab:libero_lora_performance_full}
\end{table}

\begin{table}[h]
\centering
\caption{\textbf{LoRA fine-tuning (8$\times$32 batch size)} success rates evaluated for different training steps.
500 evaluations per \libero task suite, 3 seeds. OpenVLA results copied from \cite{kim2024openvla}.}

\begin{tabular}{rccccccc}
\toprule
 & \multicolumn{5}{c}{\ourvla} & \multicolumn{1}{c}{OpenVLA} \\ 
\cmidrule(lr){2-6} \cmidrule(lr){7-7}
\libero task & 40K steps & 50K steps & 60K steps & 70K steps & 80K steps \\ 
\midrule
spatial & 84.9 \tiny{$\pm0.7$}\% & \textbf{85.7} \tiny{$\pm1.5$}\% & 85.5 \tiny{$\pm1.0$}\% & 85.7 \tiny{$\pm1.2$}\% & 84.3 \tiny{$\pm0.7$}\% & 84.7 \tiny{$\pm0.9$}\% \\ 
object & 86.9 \tiny{$\pm1.2$}\% & 87.3 \tiny{$\pm0.9$}\% & 87.6 \tiny{$\pm0.8$}\% & \textbf{88.7} \tiny{$\pm0.9$}\% & 87.3 \tiny{$\pm0.8$}\% & 88.4 \tiny{$\pm0.8$}\% \\ 
goal & 79.9 \tiny{$\pm1.6$}\% & \textbf{80.2} \tiny{$\pm1.4$}\% & 79.1 \tiny{$\pm1.3$}\% & 79.9 \tiny{$\pm0.6$}\% & 79.5 \tiny{$\pm0.7$}\% & 79.2 \tiny{$\pm1.0$}\% \\ 
10 & 55.5 \tiny{$\pm1.9$}\% & 54.2 \tiny{$\pm0.6$}\% & 55.2 \tiny{$\pm1.0$}\% & 54.5 \tiny{$\pm0.7$}\% & \textbf{55.9} \tiny{$\pm1.5$}\% & 53.7 \tiny{$\pm1.3$}\% \\ 
\midrule
average & 76.8 \tiny{$\pm0.7$}\% & 76.9 \tiny{$\pm1.0$}\% & 76.9 \tiny{$\pm0.4$}\% & \textbf{77.2} \tiny{$\pm0.8$}\% & 76.7 \tiny{$\pm0.7$}\% & 76.5 \tiny{$\pm0.6$}\% \\ 
\bottomrule
\end{tabular}
\label{tab:libero_lora_performance_full_32x8}
\end{table}

\subsection{Additional full fine-tuning results}
\label{sec:appendix_full_fine_tuning}

To determine if the higher performance of \ourvla is due to simply due to the larger batch sizes that it enables on the same hardware, or perhaps due to some other effect such as inductive bias, we performed an additional \ourvla full fine-tuning experiment with a training batch size of $8\times32$ (matching the default batch size of OpenVLA) instead of $8\times64$ (as used in Section~\ref{subsec:full_fine_tuning}). 

\begin{figure}[h] 
    \centering
    \includegraphics[width=\textwidth]{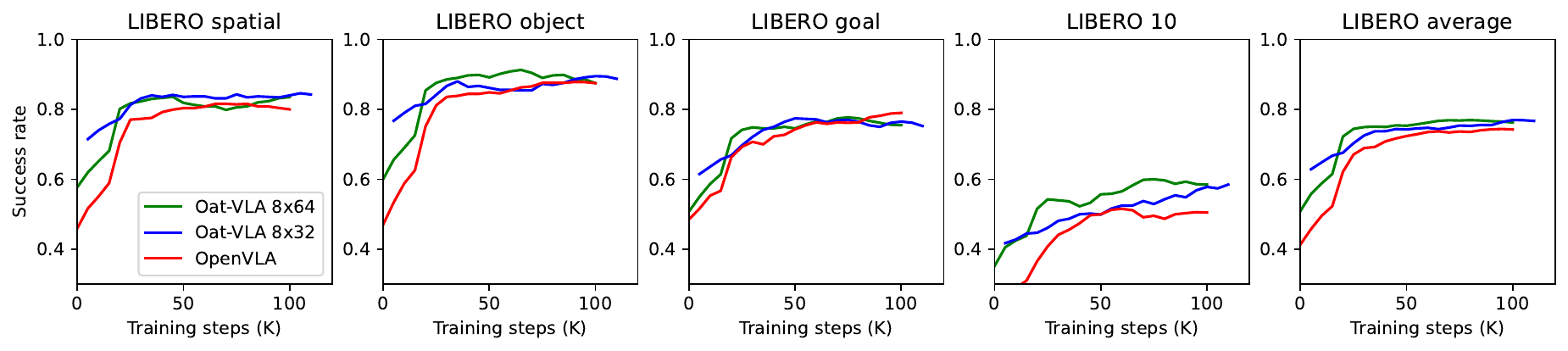}
    \includegraphics[width=\textwidth]{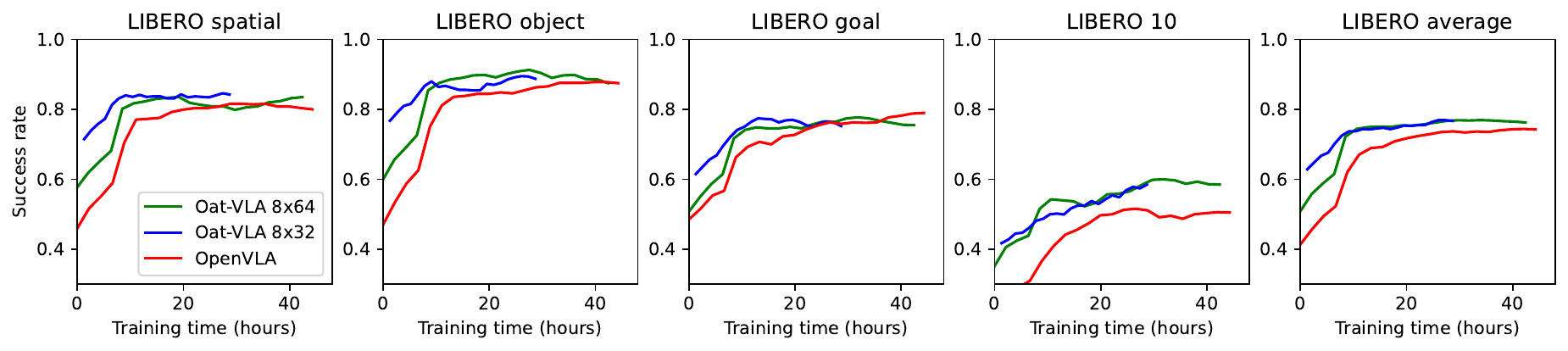}
    \caption{\textbf{Full fine-tuning} results at \textbf{$8\times64$} and \textbf{$8\times32$} batch sizes. 
    Evaluationed on LIBERO, every 5K training steps, 100 evaluations per task suite, single seed. The plots are mean filtered using one evaluation before and after.
    \textbf{Top row}: success rates on \libero task suites relative to training steps.
    \textbf{Bottom row}: success rates on \libero task suites relative to training time, as measured on an 8$\times$H100 node.}
\label{fig:libero_full_fine_tuning_8x32}
\end{figure}

Figure~\ref{fig:libero_full_fine_tuning_8x32} shows the results. We plot success rates on \libero, both relative to the number of training steps and relative to training time. We conclude that at $8\times32$ training batch size, \ourvla gets better success rates faster than at $8\times64$  but converges to around the same performance in the long run. For faster experiment turn-around, using \ourvla with a $8\times32$ appears advantageous. We also conclude that the improved success rates compared to OpenVLA are not just due to the increased batch size. 

\subsection{Training GPU memory usage and throughput}

Table~\ref{tab:gpu_info} shows the observed GPU memory usage and the throughput (number of training samples processed per second) for
both LoRA and full fine-tuning. The maximum GPU memory usage (across the 8 GPUs in a single node) determines the maximum batch size.
As can be seen, FSDP (enabled during full fine-tuning) is more balanced in terms of GPU memory usage than the LoRA implementation that OpenVLA uses,
and allows for larger batch sizes despite the larger number of learnable parameters.

\begin{table}[h!]
\caption{GPU memory usage and throughput (training samples per second) on an 8$\times$H100 node.}
\centering
\begin{tabular}{rcccccc}
\toprule
\textbf{ } & \textbf{ Fine-tuning} & \textbf{ } & \textbf{Mean GPU} & \textbf{Max GPU} & \textbf{Training samples} \\
\textbf{Model} & \textbf{Method} & \textbf{Batch size} & \textbf{memory (GiB)} & \textbf{memory (GiB)} & \textbf{per second} \\
\midrule
OpenVLA & Full & $8\times32$ & 57.6 & 61.6 & 157.4 \\
\ourvla & Full & $8\times32$ & 58.7 & 60.9 & 268.7 \\
\ourvla & Full & $8\times64$ & 78.9 & 79.6 & 320.4 \\
\midrule
OpenVLA & LoRA & $8\times16$ & 68.0 & 74.1 & 224.6 \\
\ourvla & LoRA & $8\times16$ & 40.7 & 47.1 & 307.2 \\
\ourvla & LoRA & $8\times32$ & 55.8 & 62.4 & 353.3 \\
\ourvla & LoRA & $8\times48$ & 70.9 & 75.7 & 384.0 \\
\bottomrule
\end{tabular}
\label{tab:gpu_info}
\end{table}

\subsection{Real-world experiments}
\label{sec:appendix_real_world}

\textbf{Dataset preparation.} The dataset for real-world experiments is collected via human tele-operation, using a VR set and a controller to map the human motions and intentions, e.g., closing the gripper, to the robot. After collecting the trajectories, we apply some basic operations. First, we take the sequences collected at 60 Hz, and we subsample them at 10 Hz. Then, we extract actions in the format $[\Delta x, \Delta \theta, \text{grip}]$, that is delta position, delta orientation and the observed gripper position. Finally, we filter out no-motion operations, i.e. having $\Delta x$ and $\Delta \theta$ equal to zero.

\textbf{Evaluation.} During evaluation, we map the VLA's commands to the robot as they are, with the exception of the gripper command. When the agent is closing the gripper, we close it by an additional 1-2 cms, to ensure the grasp of the object is secure. Otherwise, the agent will close the gripper barely enough to hold the object, as seen in the training dataset. 

In the following, you find a detailed description of how each in-distribution task is executed and randomized. For out-of-distribution tasks, we follow the same procedures, but we swap one of the main actors in the scene (e.g. the object to pick or the placing target).
\begin{itemize}[nolistsep]
    \item \textit{Place the banana in the green bowl.} There is a green bowl and 3 objects in the scene, a banana, a tomato and a blue hexagon, with assigned locations (see Figure \ref{fig:real_world}). We perform 5 executions with the banana in one location, and 5 executions in another location. The banana orientation can be vertical (2/5 executions per location), horizontal (2/5 executions per location) or diagonal (1/5 executions per location). 

    \item \textit{Place the red cube in the brown bag.} There is a brown bag and 3 objects in the scene, a red cube, a banana, and a lettuce leaf, with assigned locations (see Figure \ref{fig:real_world}). We perform 2 executions for the red cube in each location. The red cube orientation can be parallel to the robot's base, or tilted.

    \item \textit{Place the tomato left of the lettuce.} There is 3 objects in the scene: a lettuce leaf, a carrot and a tomato. The object's positions are randomized, around certain areas. However, the main layout stays the same. We perform 5 executions with randomized locations. Success is assigned only for the tomato being placed on the table, close to the lettuce, and left of the lettuce.

    \item \textit{Place the zucchini in front of the green cube.} There is 5 objects in the scene: a shape sorting box, a zucchini, a purple die, a rubber duck and a green cube. The object's positions are randomized, around certain areas. However, the main layout stays the same. We perform 4 executions with randomized locations. Success is assigned only for the zucchini being placed on the table, close to the green cube, and in front of the green cube.    
\end{itemize}

\subsection{Inference time, memory usage}
\label{sec:appendix_inference_time}

As stated in Section~\ref{sec:limitations}, \ourvla does not offer an advantage in terms of execution time at inference time. Here we provide measurements on how inference time scales with batch size, and a detailed break-down of inference time.

\subsubsection{Scaling of inference time with batch size}

\begin{figure}[h]
    \centering
    \includegraphics[width=0.4\textwidth]{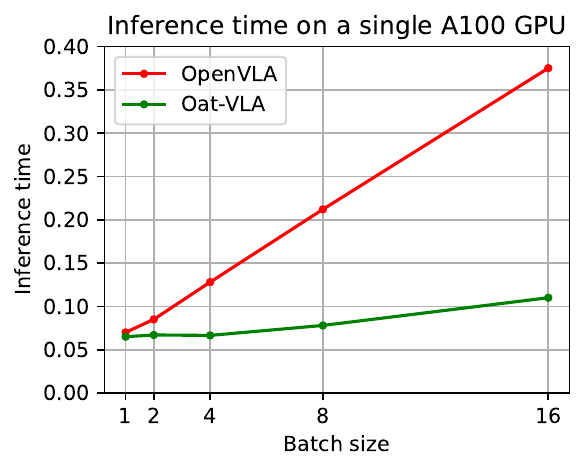}
    \caption{Inference time of the LLM, measured on a single A100 GPU, at different batch sizes.}
    \label{fig:inference_time}
\end{figure}

Figure~\ref{fig:inference_time} shows the inference time to generate a single action for various batch sizes. We measured on a single A100 GPU how long it takes the LLM to generate the action tokens from
a tokenized prompt (i.e., excluding overhead such as language tokenization and the vision back-bone).

For OpenVLA the typical number of tokens in the prompt is in the order of 300 and the inference time grows quite linearly with the batch size. For \ourvla, the number of tokens is in the order of 60 and up to a batch size of 4 the inference time is constant as it is dominated by moving the weights of the LLM from GPU RAM to GPU cache. Methods that use parallel inference (e.g., OpenVLA-OFT~\cite{kim2025fine}) might take advantage of this by increasing the batch size "for free".

\subsubsection{Detailed break-down of inference time and memory usage}

Here we provide additional data about the inference time of real-world experiments in the paper, performed on an RTX A5000. Under the prompt ``place the banana in the green bowl", the action generation with \textcolor{red}{OpenVLA} takes \textcolor{red}{284ms}, while with \textcolor{Green}{Oat-VLA} it takes \textcolor{Green}{268ms}, \textit{a reduction of 6\% in inference time}. 
During inference time, the use of key-value caching reduces the amount of computation required after the first forward pass (i.e. first action token generated). The first forward pass takes \textcolor{red}{102ms} in OpenVLA, of which \textcolor{red}{18ms} are the visual encoding process. In Oat-VLA, the first forward pass takes \textcolor{Green}{94ms}, of which visual encoding takes \textcolor{Green}{62ms}. Following forward passes (for the remaining 6 action tokens) take \textcolor{red}{29ms} for OpenVLA and \textcolor{Green}{28ms} for Oat-VLA. The memory usage for Oat-VLA is of \textcolor{Green}{17.13GB} and is higher than the \textcolor{red}{15.28GB} required for OpenVLA, due to the additional modules employed.
Thus, in addition to drastically reducing training time, Oat-VLA also slightly reduces the action prediction time, and overall inference time, at the costs of a relatively larger time and memory required for visual processing.
\vspace{0em}

\subsection{Ablation: agent-centric token grid size}

Early on during the development of \ourvla, we performed an experiment to see if the size of the agent-centric token grid matters, with results shown in  Table~\ref{tab:agent_centric_grid_size}. We found the difference insignificant enough and settled
for the smaller  option. 

\begin{table}[h!]
\centering
\caption{Success rates on \libero of an early version of \ourvla with two different agent-centric token grid sizes. 
60K training steps (batch size 8$\times$64), 100 evaluations per task suite, single seed.}
\begin{tabular}{rcc}
\toprule
\textbf{LIBERO task suite} & $3\times3$ grid & $5\times5$ grid \\
\midrule
Spatial & 77\% & 78\% \\
Object & 89\% & 90\% \\
Goal & 81\%  & 79\% \\
10 & 43\% & 44\% \\
\bottomrule
\end{tabular}
\label{tab:agent_centric_grid_size}
\end{table}

\subsection{Ablation: object-centric tokenization:  7 slots or 15 slots}

Early on during the development of \ourvla, we performed an experiment to see if the number of object-centric tokens matters, with results shown in Table~\ref{tab:object_centric_num_slots}. Again, we found the difference insignificant enough and settled
for the smaller  option. 

\begin{table}[h!]
\centering
\caption{Success rates on \libero of an early version of \ourvla with two different numbers of object-centric tokens.
60K training steps (batch size 8$\times$64), 100 evaluations per task suite, single seed.}

\begin{tabular}{rcc}
\toprule
\textbf{LIBERO task suite} & 7 tokens & 15 tokens \\
\midrule
Spatial & 82\% & 86\% \\
Object & 82\% & 86\% \\
Goal & 76\%  & 72\% \\
10 & 57\% & 40\% \\
\bottomrule
\end{tabular}
\label{tab:object_centric_num_slots}
\end{table}

\subsection{FT-Dinosaur slots as object-centric tokens}

Early on in the development of \ourvla we experimented with using an object-centric representation directly as a  way of regrouping image patch features into visual tokens. Specifically, we tried using \textit{FT-Dinosaur}~\cite{didolkar2024ftdino}, a self-supervised segmentation model built on Slot Attention~\cite{locatello2020object}. 

FT-Dinosaur learns a function \(f:\; \mathbb{R}^{N \times F_v} \rightarrow \mathbb{R}^{M \times F_s}\) to compress the original ViT (Vision Transformer) patch embeddings into a fixed set of slots, where \(N\) is the number of ViT patches, \(M\) is the (user-specified) number of slots with \(N \gg M\), and \(F_v\) and \(F_s\) are the ViT and slot feature dimensions, respectively. The model is trained with a reconstruction loss that decodes the original patch features from the slots, encouraging each slot to capture a coherent, semantically meaningful region of the scene. This yields natural segmentation while keeping the representation size constant across images, unlike typical segmentation pipelines whose output dimensionality varies with the number of masks. For this work we first train FT-Dinosaur to reconstruct \dinosiglip\ features on the Bridge~\citep{walke2023bridgedata} and LIBERO~\citep{liu2023libero} datasets and then integrate it into OpenVLA. Crucially, integration into OpenVLA is done without the FT-Dinosaur reconstruction loss.

Starting from the OpenVLA checkpoint~\cite{openvlacheckpoint} we perform full fine-tuning on the full \libero dataset. Table~\ref{tab:ftdinosaur0} summarizes the performance obtained when the patch tokens of OpenVLA are replaced by seven FT-Dinosaur slots. To put this in perspective, seven slots compress the \(16\times16\) ViT grid by a factor of \(37\). Compared with the original patch-based backbone, end-to-end slot fine-tuning narrows the gap on \textsc{Spatial}, \textsc{Object}, and \textsc{Goal} categories, but still lags far behind on \textsc{10} sequences.

\begin{table}[h]
\centering
\caption{FT-Dinosaur object-centric results. 100 evaluations per \libero task suite, single seed.
The \ourvla baseline is the evaluation of the 80K training step checkpoint in Section~\ref{subsec:full_fine_tuning}. }
\resizebox{0.8\textwidth}{!}{
\begin{tabular}{cccccc}
\toprule
 & \multicolumn{1}{c}{Frozen FT-Dinosaur} & \multicolumn{2}{c}{Training FT-Dinosaur} & \ourvla \\
 \cmidrule(lr){2-2}
\cmidrule(lr){3-4}
 & Training VLA & Frozen VLA & Training VLA  &  \\ 
\midrule
Spatial & 44\% & 43\% & 58\% & 82\% \\ 
Object  & 58\% & 53\% & 79\% & 93\% \\ 
Goal    & 44\% & 47\% & 56\% & 76\% \\ 
10    & 14\% & 9\% & 10\% & 59\% \\ 
\bottomrule
\end{tabular}
}
\label{tab:ftdinosaur0}
\end{table}

\paragraph{Qualitative behavior.}
During roll-outs in \libero the gripper consistently reaches the vicinity of the target, yet fine-grained control is unreliable: the agent often mis-grasps, bumps, or drifts off contact after a correct approach.  
We attribute these failures to three complementary factors:

\begin{itemize}[nolistsep]
    \item \textit{Lack of high-resolution cues.} With only seven slots, the policy may not “see’’ the millimeter-scale details needed for secure grasps or tight insertions.
    \item \textit{Under-constrained fine-tuning.} Jointly training FT-Dinosaur without its decoder loss occasionally destabilizes learning, suggesting that the original reconstruction objective plays a crucial regularizing role.
    \item \textit{Feature-space mismatch.} Even after fine-tuning, the slot features do not match the visual statistics of the original \dinosiglip\ patches used by OpenVLA.
\end{itemize}

These findings prompted us to start experimenting with agent-centric tokens, and at the same time we started using FT-Dinosaur only for its masks instead of for its slot features. This led to the \ourvla architecture described in Section~\ref{sec:method}.

\end{document}